\documentclass[conference,a4paper]{IEEEtran}
\let\labelindent\relax 
\usepackage[pagebackref=true,hidelinks]{hyperref}
\usepackage{times}
\usepackage{graphicx}
\usepackage{epsfig}
\usepackage{float}
\usepackage{amsmath}
\usepackage{amssymb}
\usepackage{multirow}
\usepackage{enumitem}

\newcommand{\suptensor}[1]{\mathfrak{S}^{d}}

\newcommand{\comment}[1]{}

\DeclareMathOperator{\DKL}{KL}

\newcommand{\etal}{\textit{et~al.}}

\usepackage{etoolbox}
\def\tablename{Table}
\makeatletter
\patchcmd{\@makecaption}
  {\\}
  {.\ }
  {}
  {}
\makeatother

\begin{document}
\title{Human Pose Forecasting via Deep Markov Models}

\author{
  \IEEEauthorblockN{
    Sam Toyer\IEEEauthorrefmark{1},
    Anoop Cherian\IEEEauthorrefmark{1}\IEEEauthorrefmark{2},
    Tengda Han\IEEEauthorrefmark{1},
    and Stephen Gould\IEEEauthorrefmark{1}\IEEEauthorrefmark{2}}
  \IEEEauthorblockA{
    \IEEEauthorrefmark{1}The Australian National University\\
    \IEEEauthorrefmark{2}Australian Centre for Robotic Vision\\
  {\tt\small \{sam.toyer, anoop.cherian, tengda.han, stephen.gould\}@anu.edu.au}}
}

\maketitle

\global\csname @topnum\endcsname 0
\global\csname @botnum\endcsname 0

\begin{abstract}
Human pose forecasting is an important problem in computer vision with
applications to human-robot interaction, visual surveillance, and autonomous
driving. Usually, forecasting algorithms use 3D skeleton sequences and are
trained to forecast for a few milliseconds into the future. Long-range
forecasting is challenging due to the difficulty of estimating how long a person
continues an activity. To this end, our contributions are threefold: (i) we
propose a generative framework for poses using variational autoencoders based on
Deep Markov Models (DMMs); (ii) we evaluate our pose forecasts using a pose-based
action classifier, which we argue better reflects the subjective quality of pose
forecasts than distance in coordinate space; (iii) last, for evaluation of the
new model, we introduce a 480,000-frame video dataset called \emph{Ikea
Furniture Assembly} (Ikea FA), which depicts humans repeatedly assembling and
disassembling furniture. We demonstrate promising results for our approach on
both Ikea FA and the existing NTU RGB+D dataset.
\end{abstract}
\section{Introduction}\label{sec:intro}

Deep learning methods have enabled significant advances in a variety of problems
in computer vision, including 2D human pose estimation. The impact of deep
methods on this problem has been so great that state-of-the-art accuracy on
existing pose estimation benchmarks is nearing human
performance~\cite{wei2016convolutional,newell2016stacked,tompson2015efficient}.
This has enabled researchers to look at more advanced scenarios than
single-frame pose estimation. This includes pose estimation on video
sequences~\cite{charles2016personalizing,pfister2015flowing}, real-time
estimation of poses~\cite{cao2016realtime}, multi-person pose
estimation~\cite{insafutdinov2016deepercut,iqbal2016multi}, and recently human
pose forecasting~\cite{fragkiadaki2015recurrent,jain2016structural}, which is
the central theme of this paper.

Forecasting of human poses is useful in a variety of scenarios in computer
vision and robotics, including but not limited to human-robot
interaction~\cite{koppula2016anticipating}, action
anticipation~\cite{koppula2013learning,huang2014action}, visual surveillance,
and computer graphics. For example, consider a robot designed to help a surgeon
manage their tools: it is expected that the robot forecasts the position of the
limbs of the surgeon so that it can deliver the tools in time. Pose forecasting
is also useful for proactive decision-making in autonomous driving systems and
visual surveillance.

\begin{figure}
  \begin{center}
  \includegraphics[width=0.8\linewidth]{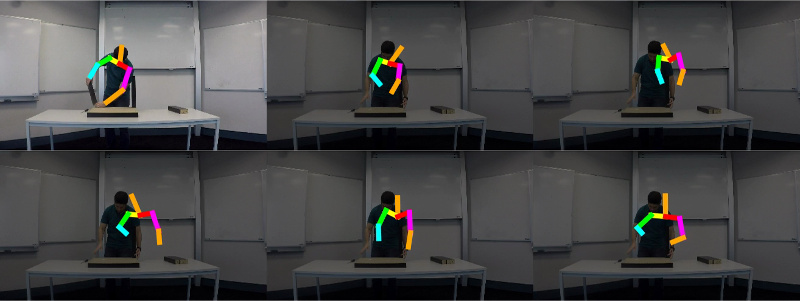}
  \end{center}
  \vspace{-1em}
  \caption{Five sampled Deep Markov Model (DMM) continuations for a sequence
    from our proposed dataset, Ikea Furniture Assembly. The top left frame shows
    the ground truth pose and image 7.5s after the beginning of the forecast
    period. The remaining frames depict the corresponding pose from each of the
    five sampled continuations.}
  \label{fig:continuations}
\end{figure}

While the problem of human motion modelling and forecasting has been explored in
the past~\cite{Han2017}, it was limited to simple scenarios such as smoothly
varying poses under periodic motions~\cite{taylor2006modeling}. As a result of
the innovations in deep methods and the availability of huge
datasets~\cite{ionescu2014human3}, this problem is beginning to be investigated
with a renewed interest. For example, Fragkiadaki~\etal{} show how to learn the
dynamics of human motion using an encoder-decoder framework for recurrent neural
networks~\cite{fragkiadaki2015recurrent}, while Jain~\etal{} present an
alternative approach which is able to incorporate high-level semantics of human
dynamics into a recurrent network via spatio-temporal
graphs~\cite{jain2016structural}. Although these newer approaches have obtained
promising results, all of them are purely deterministic models which can only
predict a single continuation of an observed sequence, and are thus unable to
account for the stochasticity inherent in human motion. Both Fragkiadaki~\etal{}
and Jain~\etal{} observe that this stochasticity can lead simple deterministic
models to produce pose sequences which rapidly converge to the mean, but they do
not attempt to resolve this issue at a fundamental level by explicitly modelling
stochasticity. A similar issue is present in the evaluation of forecasted poses:
most existing work compares forecasted poses to a single ground truth sequence,
when in reality there are often many plausible continuations.

In this paper, we propose to address both of these problems. First, to resolve
the stochasticity problem, we make use of a Deep Markov Model
(DMM)~\cite{krishnan2016structured}, which allows us to sample arbitrarily many
plausible continuations for an observed pose sequence. Second, on the evaluation
side, we use an RNN that takes a pose sequence and returns an action label. We
evaluate the quality of the forecasted pose in terms of the classification
accuracy of the evaluation RNN against the ground truth forecasted action. This
allows us to gauge the intuitive plausibility of the continued sequence without
penalising the model for choosing reasonable continuations which do not coincide
precisely with the true one.

We evaluate our proposed forecasting model on poses generated over one existing
dataset and one new one. The existing dataset is
NTU~RGB+D~\cite{shahroudy2016ntu}, which includes over 5,000,000 video frames
depicting a wide variety of actions, as well as Kinect poses for all subjects in
each frame. While NTU RGB+D is challenging, sequences in NTU RGB+D are typically
very short, and lack the kind of regularity and repetition of actions which pose
prediction mechanisms ought to be able to exploit. As such, we propose a novel
human pose forecasting dataset~\emph{Ikea Furniture Assembly}, that consists of
much longer sequences depicting 4--12 actions, each concerned with assembling a
small table. Figure~\ref{fig:continuations} shows frames from one such sequence.
While this dataset provides a simple baseline, we believe the problem of pose
forecasting is in its infancy, and that our dataset thus provides a good
platform for systematic evaluation of the nuances of the problem. Our
experiments on these datasets reveal that our proposed method demonstrates
state-of-the-art accuracy.

\begin{figure*}[ht]
  \begin{center}
  \includegraphics[width=0.7\textwidth,trim=0cm 1.5cm 0cm 1.5cm]{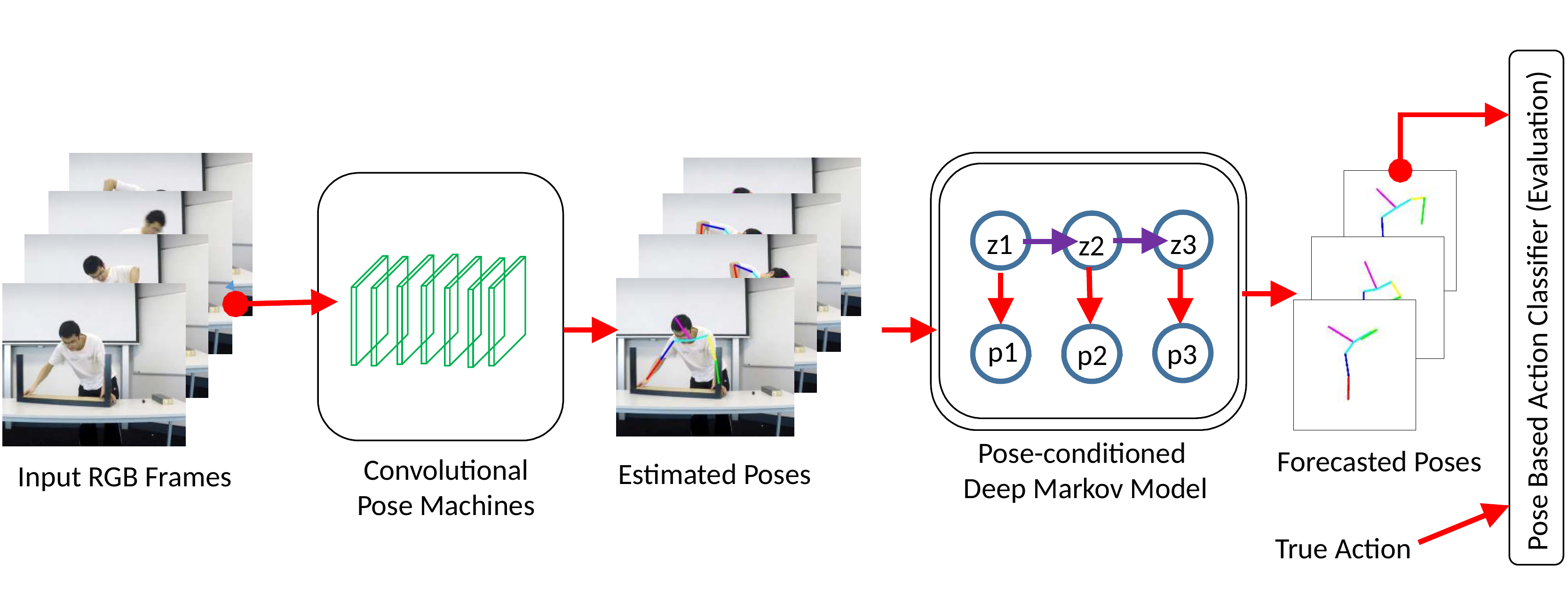}
  \end{center}
  \caption{System components and their relations.}
  \label{fig:pipeline}
\end{figure*}

\section{Background and related work}
\label{sec:related_work}

\paragraph{Pose estimation} Estimating 2D human pose from monocular images is a
well-studied problem. Approaches for static images include pictorial structure
models~\cite{yang2011articulated,chen2014articulated}, random
forests~\cite{sun2012conditional}, and coordinate~\cite{toshev2014deeppose} or
heatmap~\cite{fragkiadaki2015recurrent,wei2016convolutional,newell2016stacked}
regression with CNNs. Some approaches are able to exploit motion information by
imposing high-level graphical models~\cite{cherian2014mixing} or back-warping
joint heatmaps with flow~\cite{pfister2015flowing}; however, state-of-the-art
approaches like convolutional pose machines~\cite{wei2016convolutional} and
stacked hourglass networks~\cite{newell2016stacked} still look at only a single
frame at a time, and do not model the temporal evolution of poses.

\paragraph{Mocap modelling and synthesis} Modelling sequences of 3D motion
capture vectors is both useful in its own right---in animation, for
instance~\cite{wang2008gaussian}---and useful as a benchmark for generic
sequence learning techniques. Past approaches to this problem include Gaussian
processes~\cite{wang2008gaussian}, switching linear dynamic
systems~\cite{pavlovic2000learning} and Boltzmann
machines~\cite{sutskever2008recurrent}.

It should be noted that our work differs from the majority of existing
literature on mocap synthesis in two respects. First, we consider the problem
of 2D pose estimation in image-relative coordinates, as opposed to 3D pose
estimation in scene-relative coordinates. Second, our proposed benchmark does
not allow algorithms to make use of a ground-truth pose sequence at test time,
thereby forcing them to learn to exploit visual cues.

In the vision community, mocap modelling has sometimes been used as a motivating
application of temporally aware neural network architectures. In addition to 2D
pose estimation and forecasting, Fragkiadaki~\etal{} showcase their RNN-based
sequence learning framework by using it to extend sequences of 3D motion capture
data~\cite{fragkiadaki2015recurrent}. Similarly, Jain~\etal{} apply their
general structural RNN framework to the same task, and report aesthetically
favourable results relative to Fragkiadaki~\etal{}~\cite{jain2016structural}.
Martinez~\etal{} present a third approach specifically tailored to pose
forecasting, again reporting improvements over Fragkiadaki~\etal{} and
Jain~\etal{}, particularly in the level of error on the first few frames of a
sequence~\cite{martinez2017human}.

\paragraph{Pixel-wise prediction} Recent work in motion representation has
focused on anticipating future appearance, motion or other attributes at the
pixel level. Structured random forests~\cite{pintea2014deja}, feed-forward
CNNs~\cite{walker2015dense} and variational
autoencoders~\cite{walker2016uncertain} have all been used to predict dense
pixel trajectories from single frames. Prediction of raw pixels---as opposed to
pixel trajectories---has been attempted using ordinary feedforward networks,
GANs~\cite{vondrick2016generating,mathieu2015deep} and
RNNs~\cite{mahjourian2016geometry,finn2016unsupervised}.

Even over short time horizons, pixel-wise prediction remains a challenging task.
State-of-the-art models are capable of predicting up to a second ahead, albeit
often with significant blurring or other visual
artefacts~\cite{mathieu2015deep,finn2016unsupervised}. We hope that, by tackling
a narrower problem, pose forecasting models will be better able to learn
long-range dynamics than models for pixel-wise appearance prediction.

\paragraph{Sequence modelling with VAEs} Variational Autoencoders (VAEs) are a
recent neural-network-based approach to modeling complex, but potentially
structured, probability distributions. An ordinary regressor or autoencoder
would be trained to minimise the distance between its output and the dataset
$X$, whereas a variational autoencoder is trained to maximise the variational
lower bound $\log p(X) - \DKL(q(Z\mid X)\ \|\ p(Z \mid X)$, where $q(Z \mid X)$ is
a (learned) variational approximation to the intractable posterior $p(Z \mid X)$
over latent variables $Z$. Not only is this lower bound to the log likelihood
efficient to maximise~\cite{kingma2013auto,rezende2014stochastic}, but doing so
confers advantages over both ordinary regressors or autoencoders:
\begin{itemize}
\item VAEs are able to place a fixed prior $p(Z)$ on latent variables as part of
the optimisation process. This enables a single output $x$ to be efficiently
sampled by choosing $z \sim p(z)$, then $x \sim p(x \mid z)$.
\item In the presence of uncertainty, regressors trained to minimise $\ell_2$ error
will tend to produce the mean of all plausible outputs, which may or may not be
a plausible output itself~\cite{doersch2016tutorial}. In contrast, VAEs are able
model multimodal output distributions, thereby yielding crisper predictions.
\end{itemize}
The attractiveness of VAEs has led to a spate of new approaches to stochastic
sequence modeling, including STORNs~\cite{bayer2014learning},
VRNNs~\cite{chung2015recurrent}, SRNNs~\cite{fraccaro2016sequential} and
DMMs~\cite{krishnan2016structured}. These approaches differ in terms of the
information which their respective generators condition on between time steps,
as well as the architectures of their (approximated) inference networks. We have
chosen to use DMMs in this paper because of their strong performance in standard
benchmark tasks and their relative simplicity.
\section{Proposed method}
\label{sec:proposed_method}

In Figure~\ref{fig:pipeline}, we show the complete pose forecasting system.
First, images are passed through a pose estimation system to obtain a sequence
of poses. The observed poses are then passed to a sequence learning model to
infer a latent representation capturing the person's anticipated motion after
the final frame. This latent representation can then be extended out to an
arbitrarily long sequence of poses using the sequence learning model's
generative capabilities. Last, the sequence of forecasted poses is evaluated
by a pose-based action classifier.

\subsection{Pose estimation}

To turn an observed sequence of images into a sequence of poses, we employ Wei
et al.'s Convolutional Pose Machines (CPMs)~\cite{wei2016convolutional}.
Specifically, a four-stage CPM is used to centre detected person bounding boxes
on their subjects, followed by a six-stage model for pose estimation. To
overcome temporal instability in the estimations, the generated pose sequences
are smoothed using a weighted moving average.

\subsection{Pose forecast model}\label{sec:dmm}

The heart of our proposed system is a pose forecasting network based on Krishnan
et al.'s Deep Markov Models (DMM)~\cite{krishnan2016structured}. For the sake of
understanding how the DMM theory pertains to pose estimation, we can define the
task of pose forecasting formally: let $p_{1:T}$ denote a sequence of poses
$p_1, \ldots, p_T$ and $z_{1:T}$ denote a corresponding series of latent
variables $z_1, \ldots, z_T$. Pose forecasting is the task of sampling from the
joint distribution $p(p_{1:T}, z_{1:T})$, where $p$ is assumed to factorise
according to
\begin{equation}
  \begin{split}
  p(p_{1:T}, z_{1:T}) = &p(z_1) \, p(p_1 \mid z_1) \\ &\times \prod_{t=2}^T p(p_t \mid z_t) \, p(z_{t} \mid z_{t-1})~.
  \end{split}
\end{equation}

The fourth panel of Figure~\ref{fig:pipeline} depicts this factorisation as a
Bayesian network. From this Bayesian network, it follows that the posterior
inference distribution $p(z_{1:T} \mid p_{1:T})$ factorises to
\begin{equation}
  p(z_{1:T} \mid p_{1:T}) = p(z_1 \mid p_{1:T}) \, \prod_{t=2}^T p(z_t \mid z_{t-1}, p_{t:T})~.
\end{equation}

\subsubsection{Variational approximation}

For learning of these distributions to be tractable, we must make several
approximations. First, we can use the following variational approximations in
place of the true generative distributions $p(p_t \mid z_t)$, $p(z_1)$, and
$p(z_t \mid z_{t-1})$, respectively:
\begin{align}
  p(p_t \mid z_t; \theta)
    &= \mathcal N(\mu_E(z_t; \theta), \Sigma_E(z_t; \theta))\\
  p(z_1; \theta)
    &= \mathcal N(\mu_{T_0}, \Sigma_{T_0})\\
  p(z_t \mid z_{t-1}; \theta)
    &= \mathcal N(\mu_T(z_{t-1}; \theta), \Sigma_T(z_{t-1}; \theta))
\end{align}
The generative model parameters $\theta$ parametrise neural networks which are
able to calculate the means ($\mu_{T_0}$, $\mu_{E}$, etc.) and covariances
($\Sigma_{T_0}$, $\Sigma_E$, etc.) of the normal transition and emission
distributions ($\mathcal N(\mu_{T_0}, \Sigma_{T_0})$, etc.). Similarly, we can
replace the (intractable) posterior inference distributions $p(z_{1} \mid
p_{1:T})$ and $p(z_t \mid z_{t-1}, p_{t:T})$ with two more learnt variational
approximations expressed in terms of inference network parameters
$\phi$:\footnote{As there are many means and covariances produced by this model,
it may help to reader to know that the subscript $E$ has been used for the mean
and covariance of the emission distribution, $T$ for those of the transition
distribution (in the generative model), and $I$ for those of inference
distribution (in the inference model). $T_0$ and $I_0$ refer to the initial
latent distribution in the generative model and inference model, respectively.}
\begin{align}
  q(z_1 \mid p_{1:T}; \phi) &= \mathcal N(\mu_{I_0}(p_{1:T}; \phi), \Sigma_{I_0}(p_{1:T}; \phi))\\
  q(z_t \mid z_{t-1}, p_{t:T}; \phi) &= \mathcal N(\mu_I(z_{t-1}, p_{t:T}; \phi), \Sigma_I(z_{t-1}, p_{t:T}; \phi))
\end{align}

Parameters $\theta$ and $\phi$ can be optimised by gradient ascent to maximise
the variational lower bound $-\mathcal L(p_{1:T}; \theta, \phi)$ of Kingma et
al.~\cite{kingma2013auto}:
\begin{align}
  \begin{split}
  \log p(p_{1:T}; \theta) &\geq -\mathcal L(p_{1:T}; \theta, \phi)\\
  &= \mathbb E_{z_{1:T} \sim q_\phi(z_{1:T} \mid p_{1:T}; \phi)}[\log p(p_{1:T} \mid z_{1:T}; \theta)]\\
  &\quad\quad - \DKL[q(z_{1:T} \mid p_{1:T}; \phi) \| p(z_{1:T}; \theta)]
  \end{split}
\end{align}
Where $\DKL[f(\cdot)\|g(\cdot)]$ denotes the KL divergence between densities $f$
and $g$, and $\mathcal L(p_{1:T}; \theta, \phi)$ is shorthand for the
variational lower bound itself.

We can further factorise this variational lower bound using the conditional
dependencies identified above:
\begin{equation}
  \log p(p_{1:T} \mid z_{1:T}; \theta)
  = \sum_{t=1}^T \log(p(p_t \mid z_t; \theta))
\end{equation}
\begin{equation}
  \begin{split}
  &\DKL[q(z_{1:T} \mid p_{1:T}; \phi) \| p(z_{1:T}; \theta)]\\
  &\quad= \DKL[q(z_1 \mid p_{1:T}; \phi)\|p(z_1; \theta)]\\
  &\quad\quad + \mathbb E_{z_{1:T} \sim q(z_{1:T}) \mid p_{1:T}; \phi)} \\
  &\quad\quad\quad\quad \sum_{t=2}^T \DKL[q(z_t \mid z_{t-1}, p_{t:T}; \phi)
    \| p(z_t \mid z_{t-1}; \theta)]
  \end{split}
\end{equation}
Recall that all involved distributions are multivariate Gaussians, and so each
KL divergence has a closed form which is amenable to optimisation with
stochastic gradient descent. Further, when training with SGD, it is generally
sufficient to approximate the expectation over the $z_t$s with a single sample
from $q(z_{1:T} \mid p_{1:T})$, and it is possible to perform stochastic
backpropagation \textit{through} this sampling operation using the
reparametrisation
trick~\cite{kingma2013auto,rezende2014stochastic}.\footnote{Readers who are
completely unfamiliar with these techniques will likely appreciate the context
provided by Doersch's tutorial-style treatment of variational
autoencoders~\cite{doersch2016tutorial}.}

Taken together, these tricks make it possible to efficiently train a neural
network to optimise the DMM objective $\mathcal L(\mathcal P; \theta, \phi)$
over the full dataset $\mathcal P = \{p^{(1)}_{1:T}, \ldots, p^{(N)}_{1:T}\}$:
\begin{equation}
  \mathcal L(\mathcal P; \theta, \phi)
  = \sum_{i=1}^N \mathcal L(p^{(i)}_{1:T}; \theta, \phi)
\end{equation}

\subsubsection{Network architecture}

To calculate the various $\mu$s and $\Sigma$s required by the variational
approximation, we use a set of networks based on Krishnan et al.'s ST-LR
model~\cite{krishnan2016structured}:
\begin{description}
  \item[$\{\mu,\Sigma\}_E(z_t; \theta)$:] The mean and covariance of the emission
    Gaussian are calculated by a simple multilayer perceptron.
  \item[$\{\mu,\Sigma\}_{T_0}$:] The mean and covariance of the first latent
    vector can be learnt directly, without need for an MLP.
  \item[$\{\mu,\Sigma\}_T(z_t; \theta)$:] The transition function is
    reminiscent of a GRU: it uses $z_{t-1}$ to compute hidden activations $h_t$
    and gating activations $g_t$, then uses elementwise multiplication with the
    $g_t$ to trade off the contributions of $z_{t-1}$ and $h_t$ to the new
    hidden state. See \cite{krishnan2016structured} for details.
  \item[$\{\mu,\Sigma\}_{I_0}(z_{t-1}, p_{t:T}; \phi)$ and
    $\{\mu,\Sigma\}_I(p_{1:T}; \phi)$:] Both pairs of means and
    covariances are calculated using the same Bidirectional Recurrent Neural
    Network (BRNN), which processes a pose sequence and then yields $z_1, z_2,
    \ldots, z_T$ in order. Note that the BRNN outputs $z_t$ after processing
    actions and poses from both the future \textit{and} the past: that is, $z_t$
    is calculated from all of $p_{1:T}$, rather than just the $p_{t:T}$ and
    $z_{t-1}$ on which it is probabilistically dependent. We retain the
    $p_{t:T}$ notation in the discussion above for theoretical consistency.
\end{description}

\subsection{Classifier-evaluator}\label{sec:class-eval}

\begin{figure}
  \begin{center}
    \begin{tabular}{cccc}
      {\footnotesize Frame 0} &
      {\footnotesize Frame 15} &
      {\footnotesize Frame 30} &
      {\footnotesize Frame 45}\\
    \includegraphics[width=0.08\textwidth]{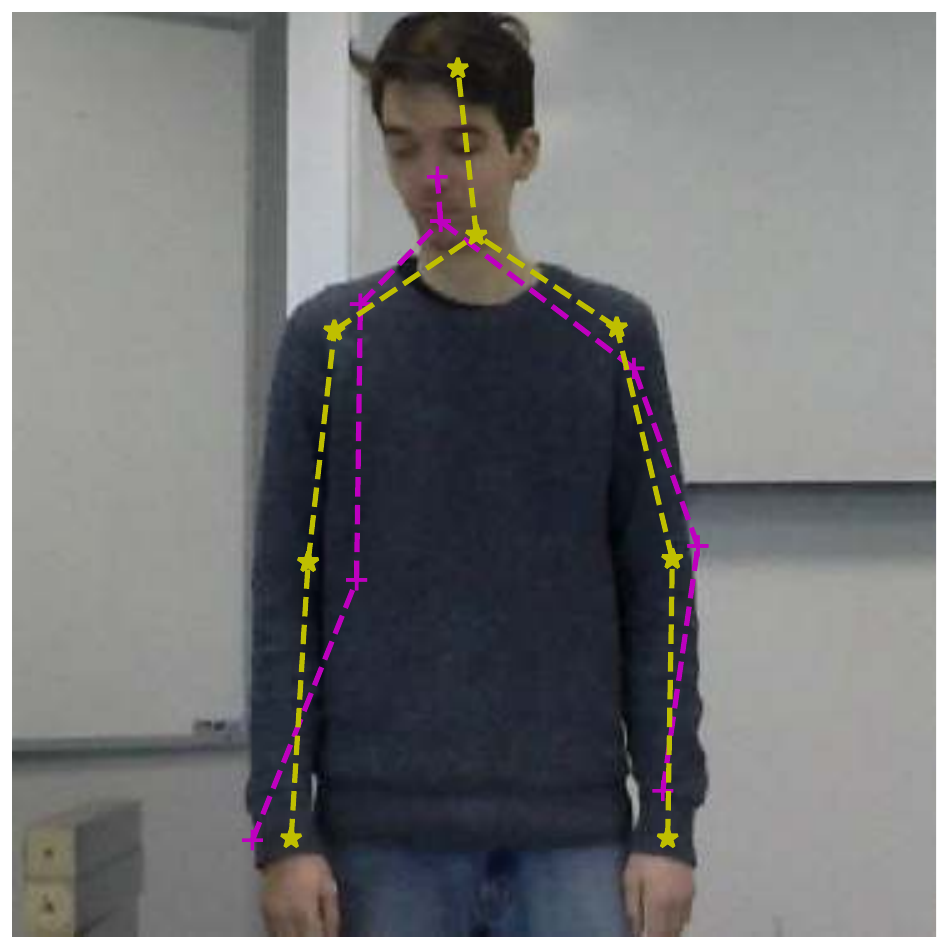} &
    \includegraphics[width=0.08\textwidth]{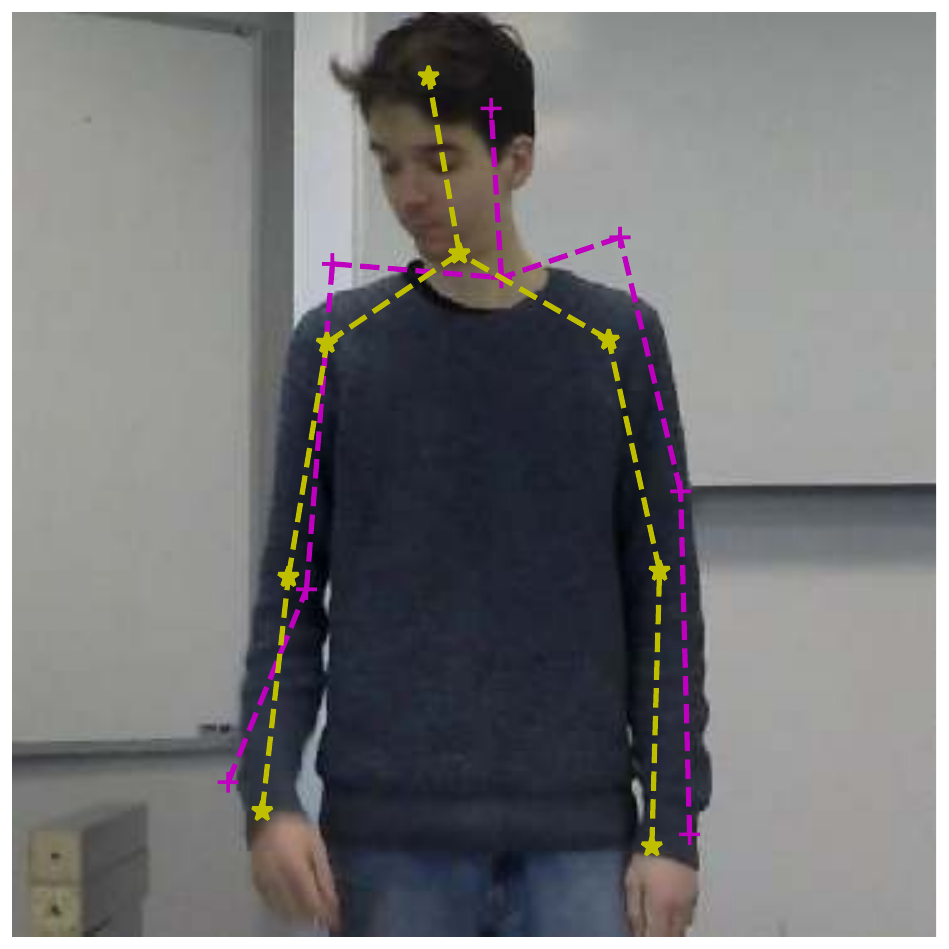} &
    \includegraphics[width=0.08\textwidth]{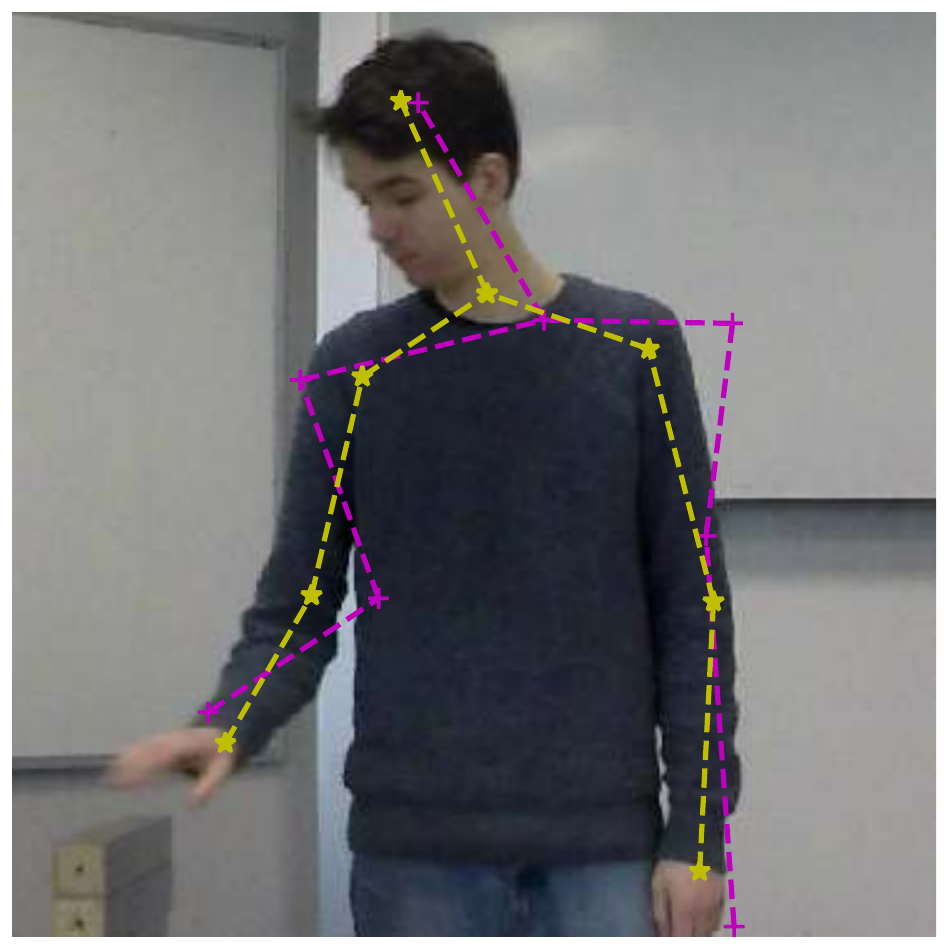} &
    \includegraphics[width=0.08\textwidth]{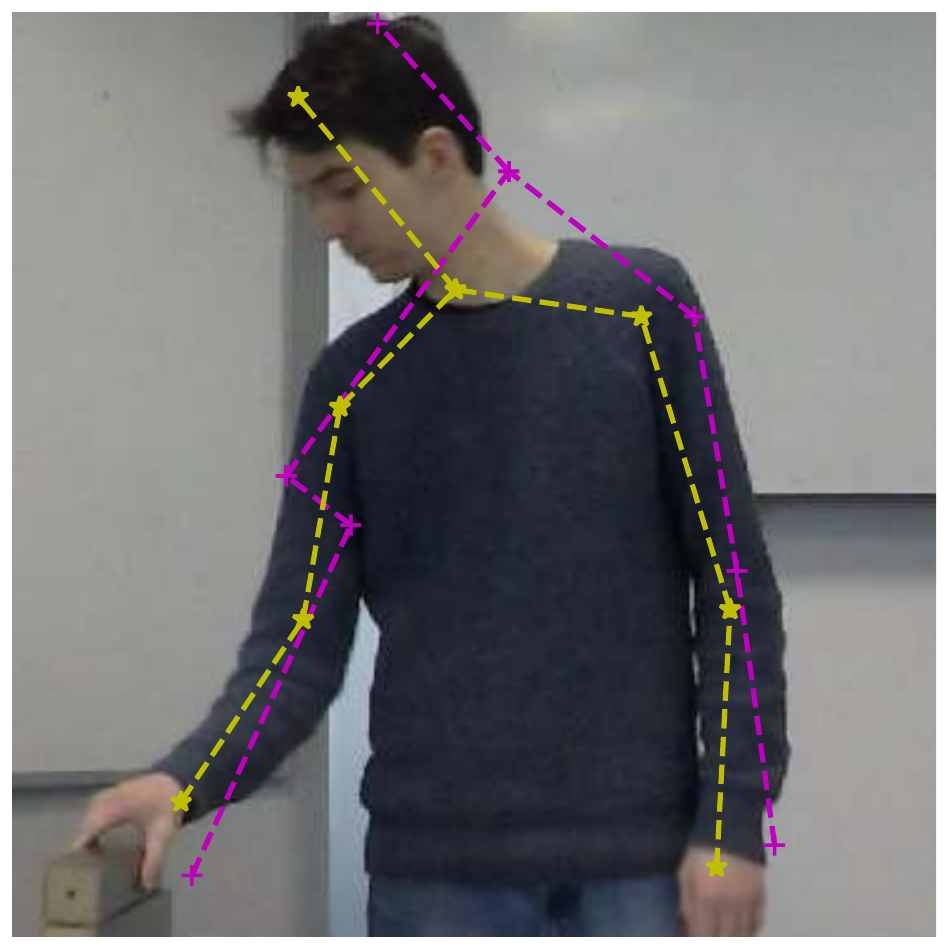}\\
      {\footnotesize Frame 60} &
      {\footnotesize Frame 75} &
      {\footnotesize Frame 90} &
      {\footnotesize Frame 105}\\
    \includegraphics[width=0.08\textwidth]{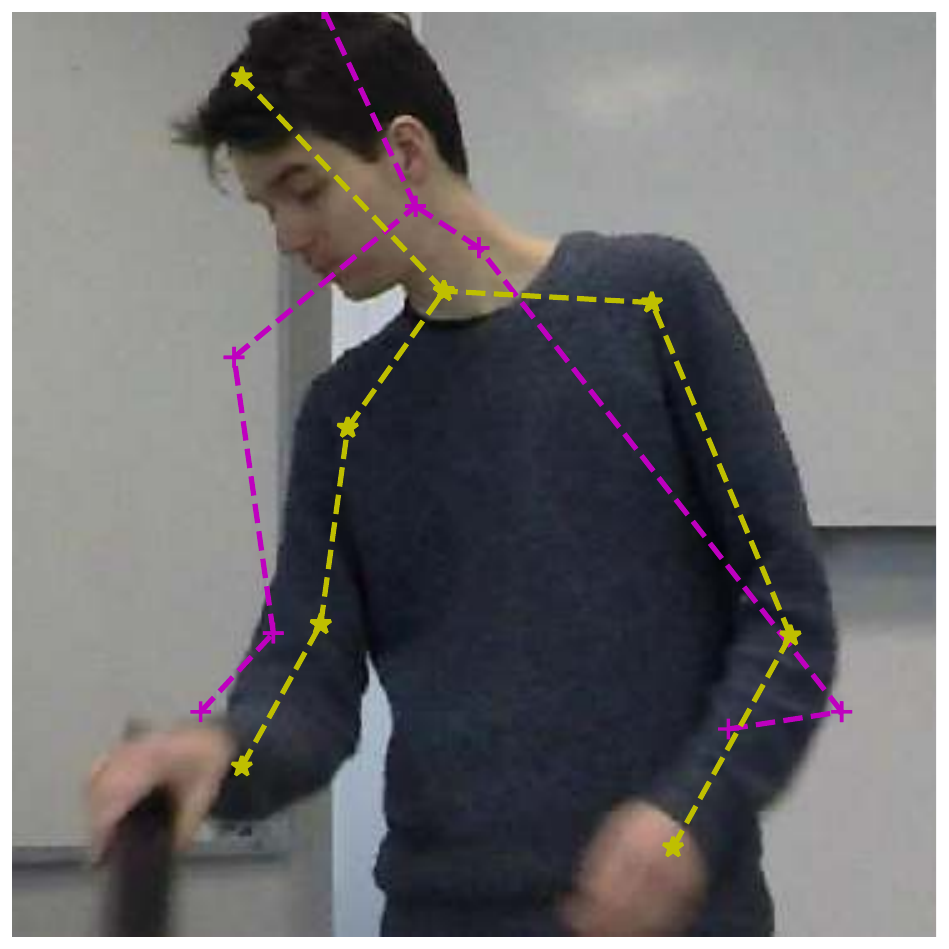} &
    \includegraphics[width=0.08\textwidth]{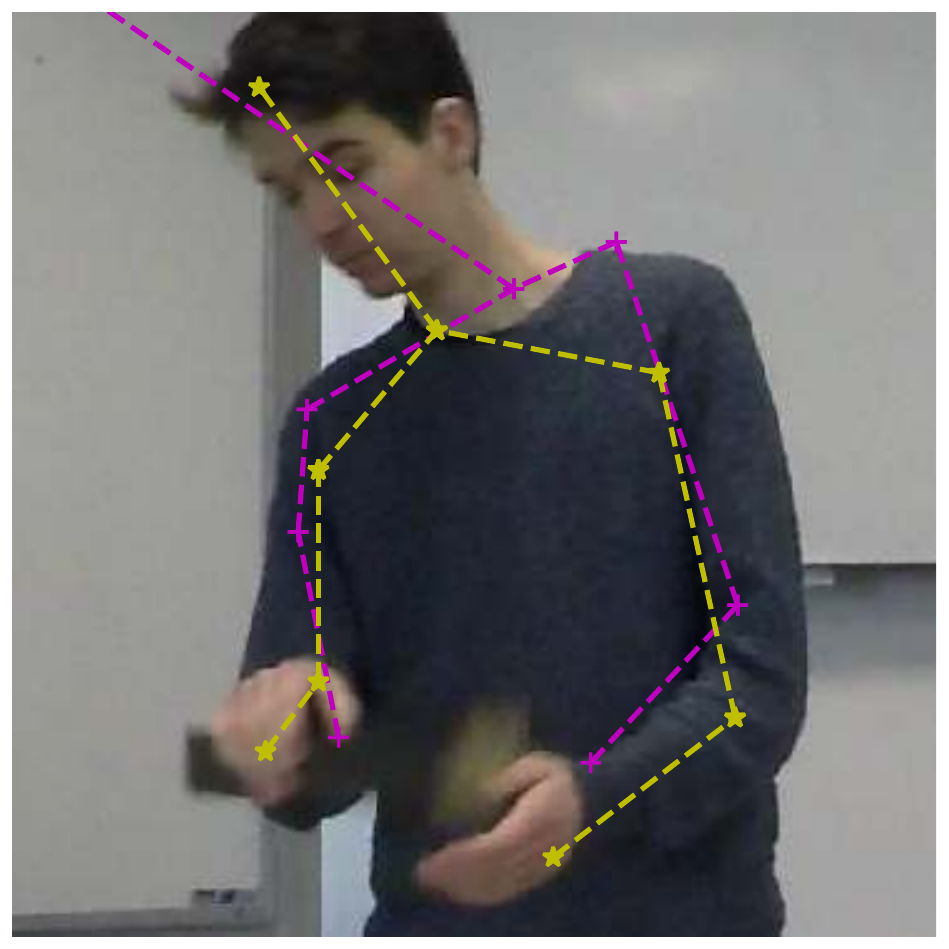} &
    \includegraphics[width=0.08\textwidth]{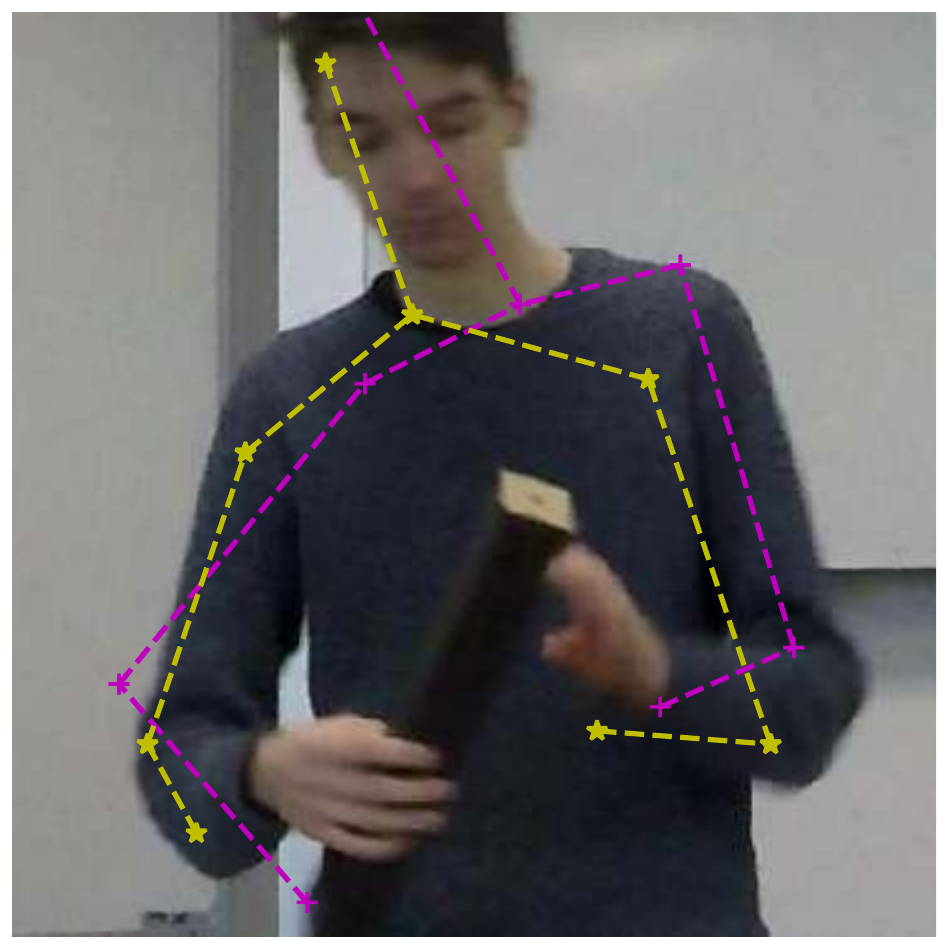} &
    \includegraphics[width=0.08\textwidth]{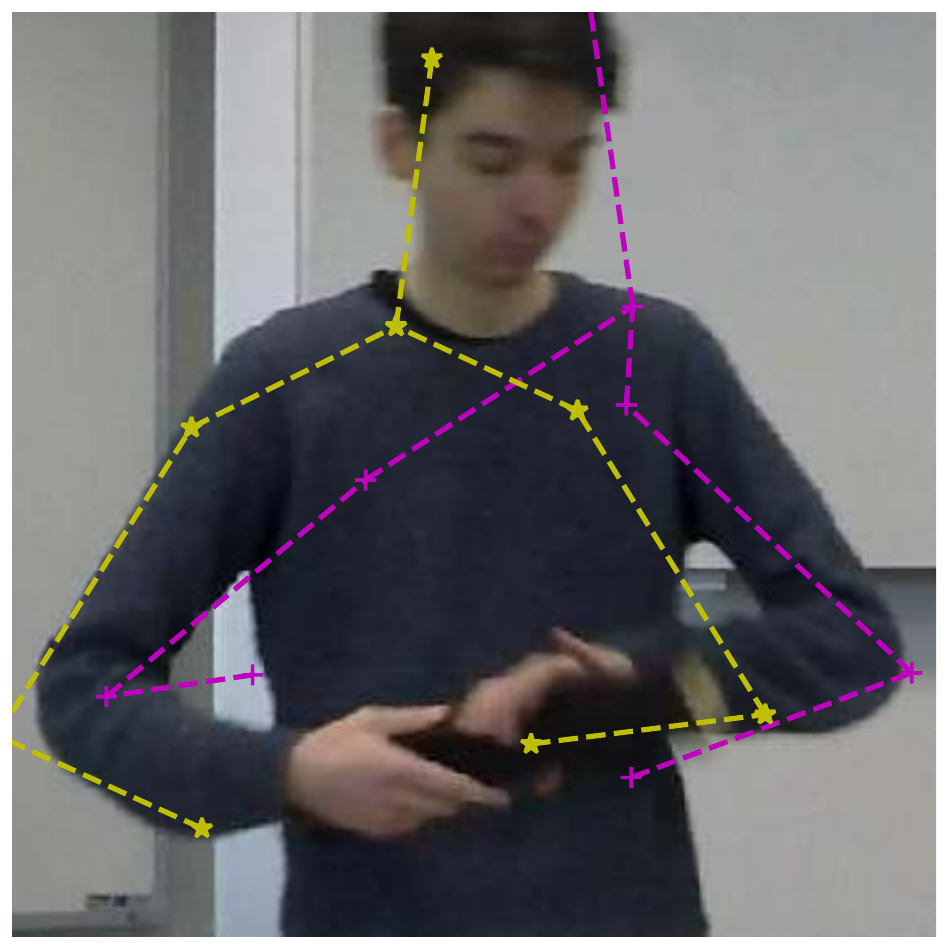}\\
    \end{tabular}\\
    \includegraphics[width=0.28\textwidth]{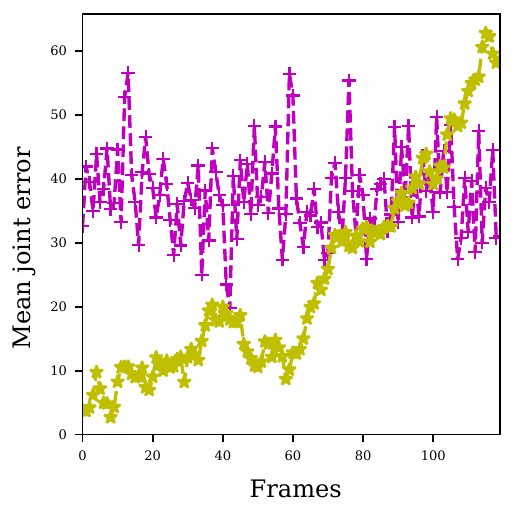}\\
  \end{center}
  \vspace{-1.5em}
  \caption{Two possible continuations of a pose sequence, and the corresponding
    $\ell_2$ distance between the continuations and the ground truth.}
  \label{fig:ext-vs-int}
\end{figure}

Evaluation of forecasted poses is difficult. The most natural approach is to
adopt a geometric measure of error and use that to compare different approaches.
For instance, Fragkiadaki et al.~\cite{fragkiadaki2015recurrent} uses a standard
location-based measure of pose estimation accuracy to report the performance of
a 2D pose forecasting system, while Jain et al.~\cite{jain2016structural}
evaluate a 3D pose forecasting system using differences in joint angles.
Unfortunately, neither of these approaches is adequate to capture human
intuition about what constitutes a natural---or even plausible---continuation of
a pose sequence.

As an example of this issue, take the two pose continuation strategies in
Figure~\ref{fig:ext-vs-int}: the magenta continuation shifts each joint of each
pose independently by a number of pixels chosen from $\mathcal N(0, 20^2)$
(again independently in both $x$ and $y$ dimensions). On the other hand, the
yellow continuation is simply the ground truth with $\mathcal N(0, 3^2)$
additional pixels of drift accumulated at each time step. The latter clearly
produces a more coherent pose sequence, but is quickly exceeded in ``accuracy''
by the former. This plainly illustrates the limitations of the $\ell_2$ distance
metric for capturing the subjective quality of pose sequence continuations in
the presence of drift. Further, although it is not illustrated in
Figure~\ref{fig:ext-vs-int}, $\ell_2$ distance can also penalise plausible, but
incorrect predictions; for instance, if a subject begins to move to the left
during a forecast was instead predicted to move to the right, it may still be
possible to produce an intuitively plausible continuation, but the continuation
would be penalised heavily by $\ell_2$ distance.

Instead of measuring $\ell_2$ distance, we score forecasted poses by checking
whether they are consistent with the action(s) they represent. This is achieved
by passing the poses through a recurrent classifier-evaluator which is trained
to recover action labels from pose sequences. The recovered action labels
associated with the forecasted pose sequences can be compared with the true
labels, providing a robust measure of the realism of the pose sequence. The
classifier-evaluator is trained using ground truth poses and action labels from
the same dataset used to train the corresponding pose forecast model(s).
However, it is \textit{not} trained adversarially, as training it in
tandem with any one forecast model would compromise its ability to provide a
model-independent measure of pose realism.
\section{Experiments}
\label{sec:expts}

\begin{table}
\begin{center}
\begin{tabular}{|l|cc|}
\hline
\multirow{2}{*}{Method} & \multicolumn{2}{c|}{Accuracy}\\
              & Ikea FA & NTU RGB+D\\\hline
Ground truth  & 91.0\%  & 83.4\%\\
Zero-velocity & 81.8\%  & 73.3\%\\
\hline
DMM           & \textbf{75.8\%}  & \textbf{70.3\%}\\
ERD           & 75.2\%  & 66.9\%\\
LSTM-3LR      & 61.9\%  & 53.5\%\\
LSTM          & 54.5\%  & 67.9\%\\
\hline
\end{tabular}
\end{center}
\caption{\textrm{\normalfont Action classifier results for both forecasted and ground-truth pose
  sequences. ``Ground truth'' shows the accuracy of the learnt classifier on the
  true pose sequences, and thus serves as an ideal for predictors to meet.}}
\label{tab:genquant}
\end{table}

\begin{figure*}
  \begin{center}
  \includegraphics{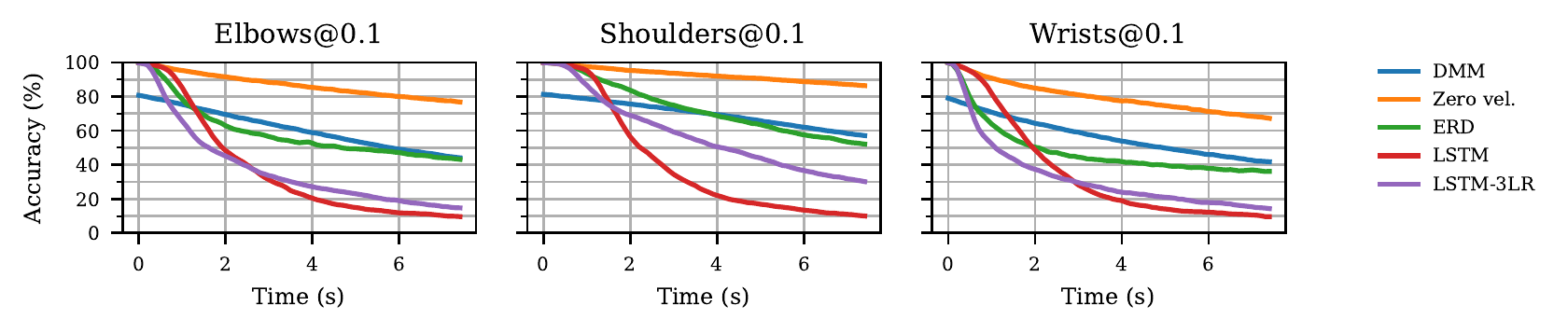}
  \end{center}
  \vspace{-2em}
  \caption{Percentage Correct Keypoints (PCK) at different times and a fixed
    threshold on Ikea FA, for a range of methods.}
  \label{fig:ikea-pck-by-time}
\end{figure*}

\begin{figure*}
  \vspace{-1em}
  \begin{center}
  \includegraphics{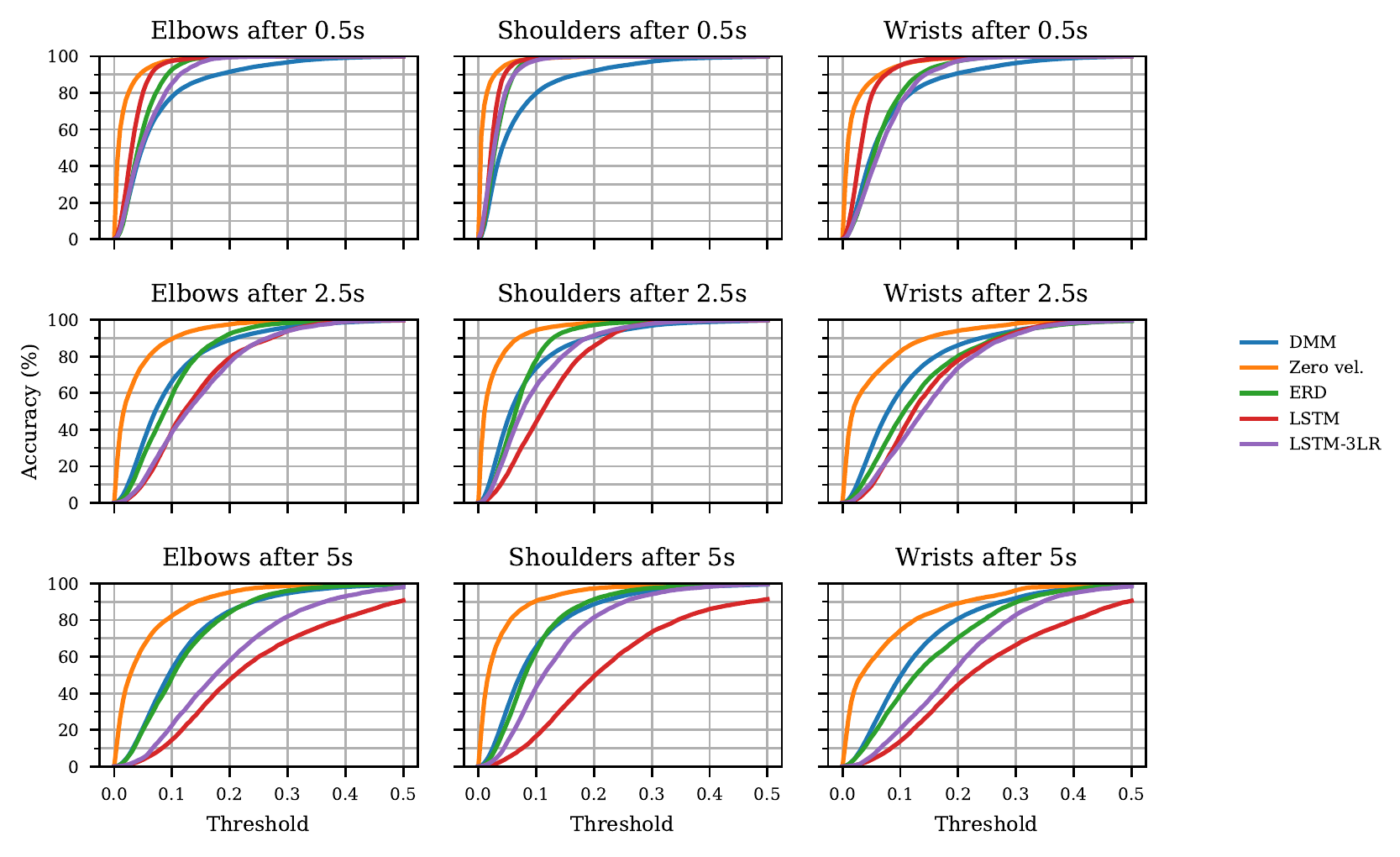}
  \end{center}
  \vspace{-2em}
  \caption{PCK at different thresholds and a handful of fixed times on Ikea FA,
    again with a range of methods.}
  \label{fig:ikea-pck-by-threshold}
\end{figure*}

\subsection{Datasets}

\paragraph{Ikea Furniture Assembly} As alluded to earlier, there are several
challenging benchmark datasets for both action recognition and pose estimation.
However, we believe these datasets are too demanding for a problem such as pose
forecasting. This is because most of the existing video benchmarks that include
both poses and actions (such as Penn Actions~\cite{wang2016mofap}, the MPII
Continuous Pose Dataset~\cite{rohrbach2012database}, or the JHMDB
dataset~\cite{Jhuang:ICCV:2013}) have either sequences that undergo significant
occlusions of body-parts, or include activities that are hard to discriminate.
As a result, it may be difficult to understand or separate the challenge imposed
by pose forecasting from that imposed by other tasks.

Towards this end, we propose a new dataset called~\emph{Ikea Furniture Assembly
} (Ikea FA) that consists of basic actions of individuals assembling a small
piece of Ikea furniture. Our goal with this dataset is to predict the pose of a
subject after a few observed frames frames containing an assembling action.
There are 101 videos in the dataset, each about 2--4 minutes long, and shot at 30
frames per second (although we downsample to 10fps in our experiments to
maximise the duration for which it is computationally feasible to predict).
There are 14 actors in the videos, and we use sequences from 11 actors for
training and validation while testing on the rest. Half of the sequences show
assembly on the floor, while the other half show assembly on a workbench. As the
frames in Figure~\ref{fig:pathologies} indicate, floor sequences tend to provide
more challenging pose variations than table ones. The dataset is available for
download on the
web.\footnote{\url{http://users.cecs.anu.edu.au/~u5568237/ikea/}}

The original Ikea FA action labels includes four ``attach leg'' actions (one for
each leg), four ``detach leg'' actions, a ``pick leg'' action, a ``flip table''
action, ``spin in'' and ``spin out'' actions, and a null action for frames which
were not labelled or could not be labelled. Since several of these actions are
indistinguishable from pose alone, we merged all attach and detach actions into
a single super-action, discarded the null actions, and merged ``spin in'' with
``spin out'', yielding only four actions.

Ikea FA does not include ground truth poses for all frames, so we used poses
estimated by CPM for our experiments. We have checked the CPM-labelled poses
against a small subset of hand-labelled poses using (strict) Percentage Correct
Parts (PCP), which measures the number of limb instances in the dataset where
both endpoints of the limb were estimated accurately to within half the true
length of the limb~\cite{eichner20122d}. By this criterion, upper arms were
localised correctly 83.0\% of the time, and lower arms 76.6\% of the time.

\paragraph{NTU~RGB+D (2D)} NTU~RGB+D~\cite{shahroudy2016ntu} is an action
recognition dataset of over 56,000~short videos, which collectively include over
4~million frames. Each sequence was recorded with a Kinect sensor, and
consequently includes RGB imagery, depth maps and 2D/3D skeletons. Each sequence
is also given a single action label from one of 60 classes, allowing us to
perform an action-classifier-based evaluation. Instead of using the full 3D
skeletons supplied with NTU~RGB+D, we limit ourselves to 2D skeletons for easier
comparison with Ikea Furniture Assembly.

Because NTU RGB+D splits each action into a separate video, most of its
sequences are only a few seconds each. Further, since the actions are largely
unrelated, it is not possible to produce meaningful sequences of actions by
stitching the videos together---one would only end up with a long sequence of
seemingly random actions that neither humans nor computers could be expected to
anticipate. However, we still wish to test the performance of our system on long
videos, so we have limited our NTU RGB+D evaluation to subsequences of 5s or
more. Unlike evaluation, training does not require a consistent sequence length,
so we still train on all sequences not used for evaluation. For both training
and evaluation, we downsample to 15fps to minimise the computational overhead of
prediction on long sequences.

As with Ikea FA, we merge NTU RGB+D's 60 actions into only seven classes for
action-based evaluation. These classes constitute ``super-actions'' which are
reasonably close in appearance. They include a super-action for a subject moving
their hand to their head, another for moving their whole body up or down on the
spot, one for walking, one for stretching their hands out in front of them,
another for kicking, a super-action for engaging in other fine manipulation with
the hands, and finally a catch-all class for actions which do not fit into the
aforementioned categories.

\paragraph{Pose parametrisation} We found that the choice of representation for
poses heavily influenced the subjective plausibility of pose sequences. Giving
the DMM and baselines an absolute $(x, y)$ coordinate for each joint resulted in
wildly implausible continuations and poor generalisation. All experiments in
this paper were instead carried out with a relative parameterisation: the
location of the head is encoded as a series of frame-to-frame displacements over
a sequence (i.e. its velocity), while the locations of each other joint was
given relative to its parent. All sequences were mean-centred and scaled to have
poses of the same height before applying this reparametrisation; after
reparametrisation, each feature was again re-scaled to have zero mean and unit
variance over the whole dataset.

\subsection{DMM and baseline configurations}

Each DMM experiment used a 50 dimensional latent space for the DMM's generative
network, and 50 dimensional state vectors for the inference network's
bidirectional recurrent units.\footnote{At evaluation time, these bidirectional
units are responsible for processing \textit{only} the observed sequence of
poses, and not the subsequent ground truth to be predicted.} The architecture is
otherwise identical to the ST-LR architecture of~\cite{krishnan2016structured}.

We compare with four baselines predictors. The first baseline is a zero-velocity
model which merely assumes that all future poses will be identical to the last
observed pose. The second is a neural network consisting of a single,
unidirectional, 128 dimensional LSTM followed by a Fully Connected (FC) output
layer. The third and fourth are Encoder-Recurrent-Decoder (ERD) and three-layer
LSTM networks (LSTM-3LR) with the same architectures as those presented
in~\cite{fragkiadaki2015recurrent}. Note that the latter two models have
significantly higher capacity than the former two: the ERD has two 500
dimensional FC layers, followed by two 1000 dimensional LSTMs, then another two
hidden 500 and 100 dimensional FC layers, before the FC output layer. Likewise,
the LSTM-3LR has a 500 dimensional hidden FC layer, followed by three 1000
dimensional LSTMs, before the FC output layer.

\begin{figure}
  \vspace{-0.8em}
  \begin{center}
  \includegraphics{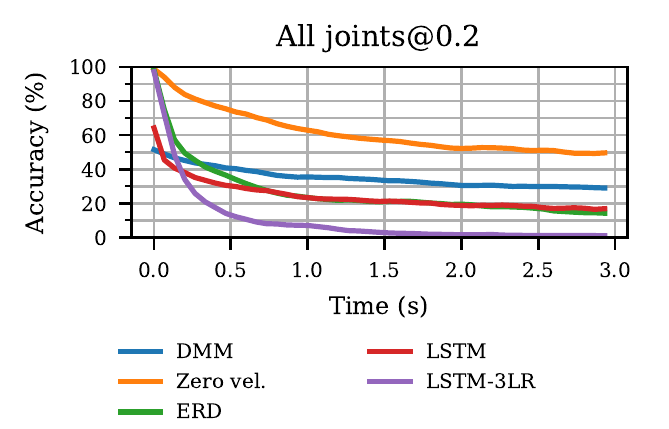}
  \end{center}
  \vspace{-1.5em}
  \caption{PCK at different times and a fixed threshold on NTU RGB+D; statistics
    for all joints have been merged together.}
  \label{fig:ntu-pck-by-time}
\end{figure}

\begin{figure}
  \vspace{-0.8em}
  \begin{center}
  \includegraphics{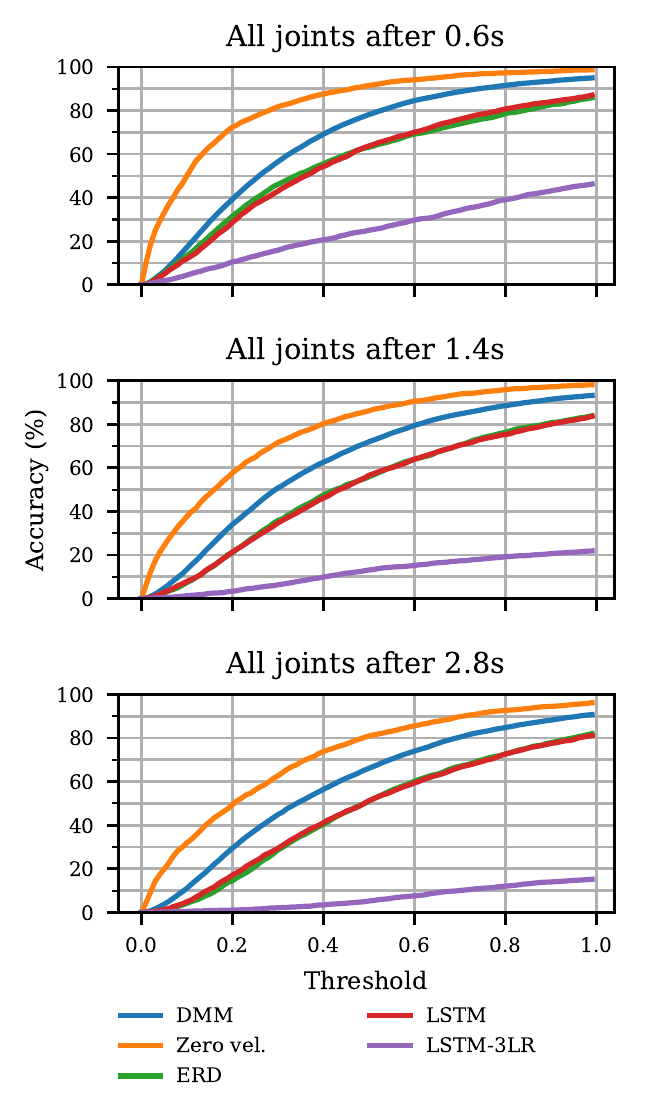}
  \end{center}
  \vspace{-1.3em}
  \caption{PCK at different thresholds and a handful of fixed times on NTU RGB+D.
    Statistics for all joints have been merged together.}
  \label{fig:ntu-pck-by-threshold}
\end{figure}

\subsection{Evaluation protocols}

\begin{figure*}
  \begin{center}
  {\small Ground truth}
  \\[0.05em]
  \includegraphics[width=0.8\textwidth]{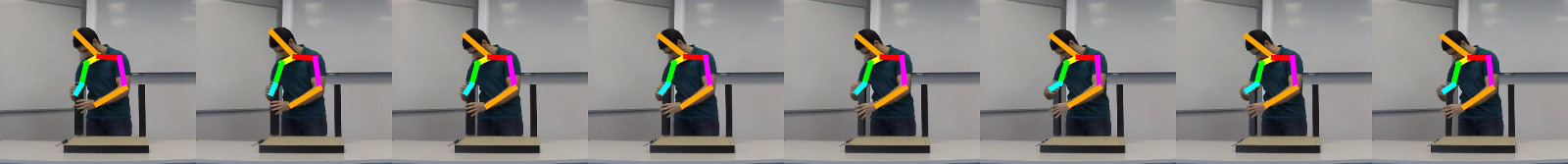}\\
  \vspace{0.1em}
  {\small DMM}
  \\[0.05em]
  \includegraphics[width=0.8\textwidth]{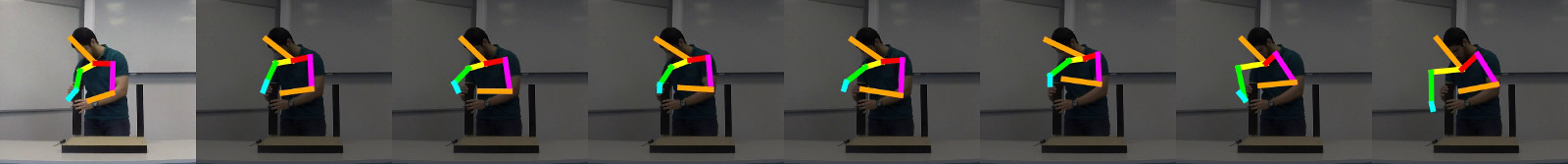}\\
  \vspace{0.1em}
  {\small ERD}
  \\[0.05em]
  \includegraphics[width=0.8\textwidth]{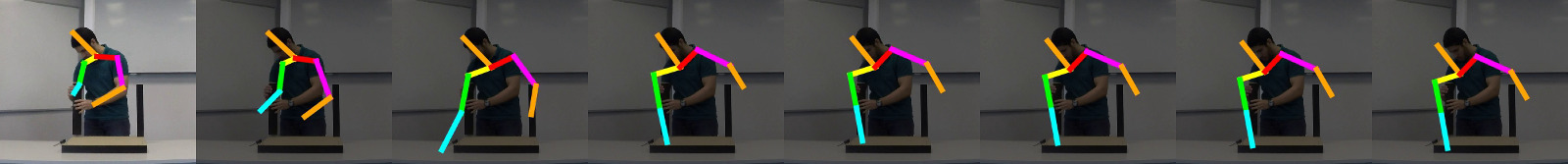}\\
  \vspace{0.1em}
  {\small LSTM}
  \\[0.05em]
  \includegraphics[width=0.8\textwidth]{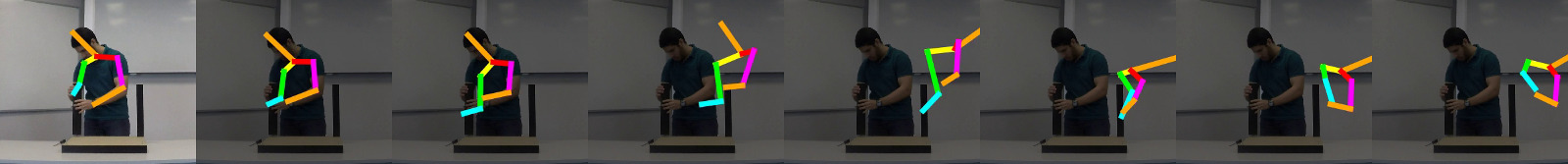}\\
  \vspace{0.1em}
  {\small LSTM-3LR}
  \\[0.05em]
  \includegraphics[width=0.8\textwidth]{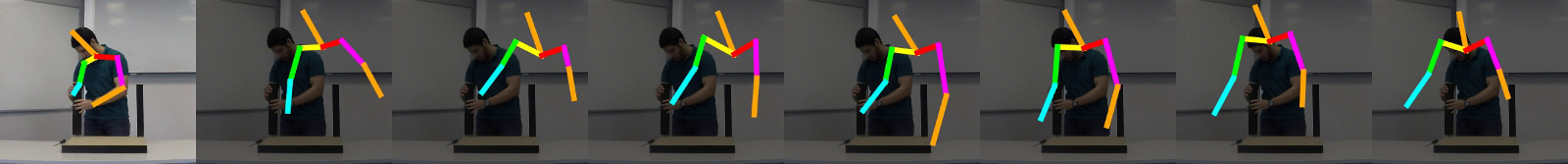}\\
  \end{center}
  \vspace{-1em}
  \caption{Qualitative comparison of four models and the ground truth over a
    7.5s forecast. Frames on the far left show the final observed poses, and
    subsequent frames show forecasted poses.}
  \label{fig:qualcomp}
\end{figure*}

Our first set of experiments focuses on displacements between forecasted poses
and the ground truth. Figure~\ref{fig:ikea-pck-by-time} and
Figure~\ref{fig:ikea-pck-by-threshold} depict the accuracy of predicted poses on
Ikea FA, while Figure~\ref{fig:ntu-pck-by-time} and
Figure~\ref{fig:ntu-pck-by-threshold} depict corresponding statistics for NTU
RGB+D. Accuracies are reported as Percentage Correct Keypoints (PCK); this
shows, for a given joint or collection of joints, the proportion of instances in
which the predicted joint was within a given distance threshold of the ground
truth. On Ikea FA, these distances are normalised to the length of the diagonal
of a tight bounding box around each pose, while on NTU RGB+D, the distances are
instead normalised using the average of displacements between a given hip (left
or right) and the opposite shoulder (right or left). Further, for the DMM, we
sampled five random continuations of each sequence and reported
\textit{expected} PCK instead of ordinary PCK; this was not necessary for the
other baselines, which are deterministic.

Our second set of experiments focuses on the degree to which forecasted pose
sequences resemble the ground truth sequence of actions for the forecast. To
recover an action sequence from a forecasted pose sequence, we apply a recurrent
classifier network consisting of a pair of 50 dimensional bidirectional GRUs,
followed by a pair of 50 dimensional FC hidden layers and an FC output layer.
Weights for this network are learnt from the training sets of each dataset. As
with the first set of experiments, we averaged the DMM's performance over five
sampled continuations per pose sequence. Table~\ref{tab:genquant} shows the
results of these experiments.

\begin{figure}
  \vspace{-0.8em}
  \begin{center}
  \includegraphics[height=0.18\linewidth]{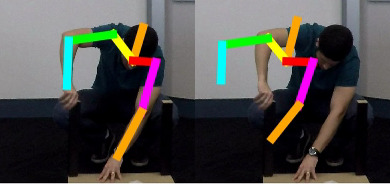}
  \includegraphics[height=0.18\linewidth]{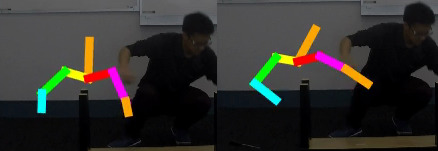}
  \end{center}
  \vspace{-1.3em}
  \caption{Typical DMM errors: on the left, the velocity-based parameterisation
    has led to drift between the original, observed pose (far left) and the
    corresponding pose recovered from the DMM (second from left). On the right,
    the DMM has made an implausible transition between two independently plausible
    poses in adjacent frames.}
  \label{fig:pathologies}
\end{figure}

\section{Discussion}
\label{sec:discuss}

The joint-position-based evaluations on Ikea FA and NTU RGB+D show that the DMM
performs best on longer forecasting horizons of several seconds or more.
Further, Table~\ref{tab:genquant} shows that we also improve on the three
``smart'' baselines (ERD, LSTM, LSTM-3LR) in terms of the consistency of our
produced poses with the ground truth action sequence. Qualitatively, we found
that the ERD, LSTM and LSTM-3LR baselines tend to start out with little error,
but quickly either converge to a mean pose or diverge to an implausible set of
poses. In contrast, the individual plausibility of poses produced by the DMM
tends not to decrease over time, and the DMM's output does not rapidly converge
to an obvious fixed point. Forecasts for each model on a single example sequence
are included in Figure~\ref{fig:qualcomp}.

One surprising---and troubling---outcome of our experiments is the high
performance of the zero velocity model. Not only does it dominate all baselines
in both types of experiment, but it also beats our DMM model. This kind of
performance has been observed by other authors, as well: in introducing the ERD,
Fragkiadaki~\etal reported that their model did not consistently outperform a
zero-velocity baseline on forecasting of upper-body
joints~\cite{fragkiadaki2015recurrent}, and more recent work shows that this is
also true of other ``state-of-the-art'' models~\cite{martinez2017human}.

There are some factors specific to the DMM which may be causing it to fall short
of the zero-velocity baseline. In large part, we expect that its inaccuracy is
due to drift in the predicted pose: as noted in Section~\ref{sec:class-eval},
small errors in the predicted centre-of-mass of a person can rapidly build up to
destroy prediction accuracy, even when the motion of limbs relative to one
another is small. This problem is exacerbated by our choice of a velocity-based
parametrisation for head location, which leads to very rapid build-up of error
during fast motions. Even when the DMM is able to observe ground truth poses,
before beginning to forecast, it still accumulates error as the velocity-based
input features do not provide the necessary information for it to correct the
drift which it accumulates. The left side of Figure~\ref{fig:pathologies}
illustrates the issue.

As Figure~\ref{fig:continuations} demonstrates, the DMM is able to produce a
good diversity of continuations for a given pose sequence. To some extent,
though, this diversity comes at the cost of temporal consistency: during
evaluation, we found that the DMM would sometimes flip between poses which were
independently plausible, but not temporally consistent, as demonstrated on the
right side of Figure~\ref{fig:pathologies}. In contrast, the deterministic
baselines offer much smoother continuations, but at the cost of poorer accuracy
and a rapid drift toward implausible poses. In principle, it ought to be
possible to obtain the best of both methods by adding a temporal smoothness
penalty to the DMM, although we have so far been unable to improve results this
way.

\section{Conclusion}
\label{sec:conclude}

We have proposed a novel application of Deep Markov Models (DMMs) to human pose
prediction, and shown that they are better able to make long-range pose
forecasts than state-of-the-art models. Further, we have introduced a new action
recognition and pose forecasting dataset called Ikea Furniture Assembly, and
proposed a mechanism for action-based evaluation of pose forecasts. Given the
inherent difficulty of making long-range motion forecasts from skeletons alone,
we believe that the most fertile ground for future research lies in the use of
visual context to enable more meaningful predictions over horizons of several
seconds and beyond.

{\small
\bibliographystyle{IEEEtran}
\bibliography{IEEEabrv,posepred_bib}}

\end{document}